%% file: main.tex
\documentclass[11pt]{article}
\ifdefined\XeTeXversion\else\pdfoutput=1\fi

\usepackage[margin=1.35in]{geometry}
\usepackage[T1]{fontenc}
\usepackage[utf8]{inputenc}
\usepackage{newtxtext,newtxmath}
\usepackage{microtype}
\usepackage{booktabs}
\usepackage{multirow}
\usepackage{graphicx}
\usepackage{amsmath}   
\usepackage{xcolor}
\usepackage[colorlinks=true,linkcolor=blue!60!black,citecolor=blue!60!black,urlcolor=blue!60!black]{hyperref}
\usepackage[numbers,sort&compress]{natbib}
\usepackage{caption}
\usepackage{pgfplots}
\pgfplotsset{compat=1.17}
\tikzset{
  col1/.style={color=blue!70!black},
  col2/.style={color=red!70!black},
  layer0/.style={color=blue!75!black},
  layer1/.style={color=cyan!55!black},
  layer2/.style={color=teal!70!black},
  layer3/.style={color=orange!85!black},
  layer4/.style={color=red!70!black},
}

\newcommand{\minima}{\textsc{Minima}}

\newcommand{\relS}{\mathrm{relS}}

\title{\bf Why Gated DeltaNet Survives 4-Bit Quantization:\\
NVFP4 W4A4 for the Recurrent Half of a Hybrid 27B LLM}

\author{Sergii~Kozyrev\\
  \texttt{sergii@mnma.ai}
  \and
  Davyd~Maiboroda\\
  \texttt{david@mnma.ai}}

\date{}

\begin{document}
\maketitle

\begin{abstract}
Hybrid large language models pair softmax attention with linear-attention
layers such as Gated DeltaNet (GDN), whose recurrent state summarizes the
entire context in fixed size. Early community 4-bit quantizations of
Qwen3.8-27B (48 GDN layers, 16 attention layers) uniformly left the GDN
block in 8- or 16-bit precision --- in particular its decay and
write-strength gate projections --- on the intuition that errors in a
recurrence accumulate over long contexts. We test that intuition by
building \minima{}: NVFP4 W4A4 on \emph{all} 496 linear layers, GDN
included. Across perplexity at 4K/32K, MMLU-Pro, GSM8K, AIME'25,
GPQA-Diamond, LiveCodeBench, and RULER retrieval to 64K, \minima{}
matches BF16 within seed noise (5-task average $-0.52$) while being the
smallest (17.5\,GiB) and fastest-prefill ($+14$--$19\%$) of the recipes
we compare, and its 32K perplexity gap \emph{shrinks} with position in
the context window. We then explain \emph{why} with a four-part
mechanism study on captured activations: (i) GDN inputs share the
residual stream's extreme outliers, but NVFP4's 16-element block scaling
localizes them, equalizing activation error across all layer roles;
(ii) the supposedly fragile gate projections are the \emph{least}
sensitive --- their softplus/exponential and sigmoid parameterizations
compress a ${\sim}11\%$ GEMM error to a ${\sim}2\%$ output error;
(iii) the delta-rule recurrence bounds injected quantization noise at a
flat plateau over 32K tokens and forgets a state impulse within
hundreds of steps, far faster than its decay-gate horizons, because
each write overwrites the state along the current key direction;
(iv) end-to-end, the per-token quantization cost washes out with
context instead of compounding. Along the way we identify and repair a
global-scale mismatch that arises when per-module-calibrated NVFP4
checkpoints are served by kernels that fuse those modules into one
GEMM, and we show that calibrated FP8 KV-cache scales are
performance-free and recover 83\% of the quantized model's long-context
KV penalty. The result is a practical recipe --- quantize everything,
ship KV scales --- and a mechanistic account of why the recurrent half
of a hybrid LLM is the \emph{easy} half to quantize. The quantized
checkpoint is released at
\url{https://huggingface.co/minima-ai/mnma_qwen3.8_27b_nvfp4}.
\end{abstract}

\section{Introduction}\label{sec:intro}

Serving cost has made 4-bit weights-and-activations (W4A4) inference a
practical target: NVFP4 pairs E2M1 4-bit values with an E4M3 scale per
16-element block and runs natively on current accelerators. At the same
time, frontier open models are increasingly \emph{hybrid}: most layers
replace softmax attention with linear-attention operators whose state is
a fixed-size matrix updated recurrently --- in Qwen3.8-27B
\citep{qwen2026qwen38}, 48 of 64 layers are Gated DeltaNet
\citep{yang2025gateddeltanet} and only 16 are full attention.

These two trends have only just begun to meet. Every public 4-bit build of Qwen3.8-27B
we examined at the outset --- and the model authors' own FP8 release --- quantizes the
MLPs aggressively but protects the GDN block: its query/key/value,
output, and especially its gate projections ($a$, controlling the decay
$\alpha_t$, and $b$, controlling the write strength $\beta_t$) stay in
8- or 16-bit. The implicit argument is natural: a recurrence
$S_t = \alpha_t S_{t-1} + \beta_t\,k_t(v_t - S_{t-1}^{\top}k_t)^{\top}$
carries its state across tens of thousands of tokens, so a per-step
quantization error should compound, and errors in the \emph{gates}
should compound fastest.

This paper shows the intuition is backwards for this architecture, and
explains why. Our contributions:

\begin{enumerate}
  \item \textbf{A true W4A4-GDN model, evaluated seriously.}
  \minima{} quantizes all 496 linear layers of Qwen3.8-27B to NVFP4
  W4A4, GDN gates included. Under a fixed serving regime (FP8 KV cache,
  identical harness, per-sample validity checks), it matches BF16 and
  two community recipes within seed noise on six accuracy suites while
  being the smallest and the fastest at prefill
  (\S\ref{sec:results}).
  \item \textbf{A mechanism study of why it works}
  (\S\ref{sec:mechanism}): activation statistics on captured 32K-token
  inputs, per-projection sensitivity replay, error propagation through
  the recurrence in lockstep FP32, and a positional decomposition of
  the perplexity gap. The chain --- block scaling localizes outliers,
  gate nonlinearities squash what remains, the delta rule actively
  erases state error, and the end-to-end gap shrinks with context ---
  turns ``it happens to work'' into an architectural account, and shows
  the protected projections were precisely the safest ones.
  \item \textbf{Serving-stack findings required to measure any of this
  correctly} (\S\ref{sec:systems}): a global-scale mismatch between
  per-module NVFP4 calibration and fused serving GEMMs that silently
  corrupts the GDN gates (and fakes \emph{better} long-context
  perplexity); a multimodal-composite serving path that degrades
  long-context scoring; and a chat-template pitfall that invalidates
  raw-completion harnesses for thinking models.
  \item \textbf{A KV-cache recipe} (\S\ref{sec:kv}): FP8 KV halves KV
  memory and moves no task score; its one visible cost --- a $+0.41$
  perplexity penalty at 32K, $3\times$ larger for the quantized model
  --- is eliminated by calibrated per-layer scales that are free at
  serving time (83\% recovered, throughput unchanged within 0.4\%).
\end{enumerate}


\section{Background}\label{sec:background}

\paragraph{Gated DeltaNet in Qwen3.8-27B.}
The model interleaves 48 GDN layers with 16 full-attention layers
(hidden size 5120). A GDN layer projects the residual stream through
four linear maps: \texttt{in\_proj\_qkv} (whose output passes a
depthwise causal convolution and SiLU before splitting into $q_t, k_t,
v_t$), an output gate \texttt{in\_proj\_z}, and two scalar-per-head gate
projections \texttt{in\_proj\_a} and \texttt{in\_proj\_b}. The gates are
parameterized in log space,
\begin{equation}
g_t \;=\; -\exp(A_{\log})\,\operatorname{softplus}(a_t + \mathrm{dt\_bias}),
\qquad
\alpha_t = e^{g_t}\in(0,1),
\qquad
\beta_t = \sigma(b_t)\in(0,1),
\label{eq:gates}
\end{equation}
and the per-head state $S_t \in \mathbb{R}^{K\times V}$ ($K{=}V{=}128$)
evolves by the gated delta rule over $\ell_2$-normalized keys and
queries:
\begin{equation}
S_t \;=\; \alpha_t\,S_{t-1} \;+\; \beta_t\,
k_t\bigl(v_t - S_{t-1}^{\top}k_t\bigr)^{\!\top},
\qquad
o_t \;=\; S_t^{\top}\,\bigl(q_t/\sqrt{K}\bigr).
\label{eq:recurrence}
\end{equation}
$\alpha_t$ is a per-token forget gate; $\beta_t$ scales a
\emph{correction}: the write replaces what the state currently predicts
for key $k_t$ by $v_t$, rather than accumulating $v_t$ blindly. The
output is gated (RMSNorm modulated by $z$) and mixed back by
\texttt{out\_proj}. Five weight matrices per layer are therefore
candidates for quantization; the community consensus protects $a$ and
$b$ entirely and keeps the rest at 8~bits.

\paragraph{NVFP4.}
NVFP4 stores values in E2M1 (4~bits) with one E4M3 scale per 16-element
block (set to $\mathrm{blockmax}/6$) and one FP32 scale per tensor.
W4A4 means both weights and activations are quantized at this
granularity, so the GEMM runs on native 4-bit tensor cores. Two
consequences matter later: a block's largest value fixes its scale, so
an outlier degrades only its own 15 neighbors; and within a block
$\max/\mathrm{RMS} \le \sqrt{16} = 4$, which bounds how ``one-hot'' a
block can be.

\section{Experimental setup}\label{sec:setup}

\paragraph{Checkpoints.}
All models are served text-only (\S\ref{sec:systems}) with vLLM 0.27.1
\citep{kwon2023vllm}, TP=1, on one RTX PRO 6000 (96\,GB, SM120, native
NVFP4).
\textbf{BF16} is the unquantized reference, extracted from the
multimodal composite. \textbf{\minima{}} (ours) applies llm-compressor
NVFP4 W4A4 to every linear layer --- 240 GDN, 64 attention, 192 MLP
projections; 496 in total --- excluding only \texttt{lm\_head},
embeddings, convolutions, and norms; it is calibrated on a frozen
128-sample $\times$ 32K-token set and served after the global-scale
harmonization of \S\ref{sec:systems}. \textbf{Unsloth} (Dynamic~v3)
and \textbf{RadixArk} (ModelOpt) are the two public NVFP4
checkpoints: both keep GDN and attention at FP8 W8A8 with $a$/$b$ in
BF16, and quantize only MLPs to NVFP4. \textbf{\minima{}+scales} is
\minima{} plus calibrated FP8 KV-cache scales (\S\ref{sec:kv}),
identical in every other tensor; it is the released
checkpoint.\footnote{\url{https://huggingface.co/minima-ai/mnma_qwen3.8_27b_nvfp4}}

\paragraph{Serving regime and harness.}
One regime for every number in this paper: FP8 KV cache
(\S\ref{sec:kv} ablates it), GPU utilization 0.85, 32K generation cap.
Accuracy suites: WikiText-2 perplexity measured at 4K and
\emph{inside} a single 32K request; MMLU-Pro and GSM8K via a
chat-template harness with thinking disabled (\S\ref{sec:systems}
explains why raw-completion harnesses are invalid for this model);
AIME'25 and GPQA-Diamond as pass@1 over four seeds (temperature 0.6,
top-p 0.95); LiveCodeBench v6 unit-test graded; RULER \citep{hsieh2024ruler} NIAH
single and multikey at 32K and 64K. Every task run passes per-sample validity
gates (empty/unextracted answers, leaked \texttt{<think>} blocks), and
truncation statistics are reported alongside scores.

\section{Main results}\label{sec:results}

\begin{table}[t]
\centering
\caption{Four models, one regime (FP8 KV, vLLM 0.27.1, TP=1, one
RTX~PRO~6000). Accuracies in \%; AIME'25 and GPQA-D are pass@1 over
the same 4 seeds for every model (per-seed scores for BF16/\minima{} in
parentheses), LCB~v6 single seed;
5-task avg = mean of MMLU-Pro, GSM8K, AIME'25, GPQA-D, LCB. RULER
(100 for all four at 32K/64K, single and multikey) is omitted. PPL@32K
is measured inside a 32K request. No pair of models is CI-separated on
any task. Decode = 1024-in/1024-out at concurrency 32; TTFT = one
32K-token prefill.}
\label{tab:main}
\small
\setlength{\tabcolsep}{4pt}
\begin{tabular}{lcccc}
\toprule
 & BF16 & \textbf{\minima{}} & Unsloth & RadixArk \\
\midrule
PPL @4K / @32K $\downarrow$ & 6.95 / 10.35 & 7.67 / 10.84 & 7.16 / 9.91 & 7.35 / 9.95 \\
MMLU-Pro & 80.4 & 79.7 & 78.9 & 79.1 \\
GSM8K & 95.5 & 95.5 & 95.4 & 95.7 \\
AIME'25 & 86.7 {\scriptsize(.83/.93/.83/.87)} & 86.7 {\scriptsize(.87/.87/.87/.87)} & 87.5 & 84.2 \\
GPQA-Diamond & 86.5 & 85.1 & 85.0 & 85.4 \\
LiveCodeBench v6 & 79.0 & 78.5 & 79.9 & 79.6 \\
5-task avg / $\Delta$ & 85.62 & 85.10 / $-0.52$ & 85.34 / $-0.28$ & 84.80 / $-0.82$ \\
\midrule
Weights in VRAM & 50.13 GiB & \textbf{17.53 GiB} & 20.23 GiB & 18.83 GiB \\
Decode tok/s @32 & 621 & 1,154 & 1,132 & \textbf{1,174} \\
TTFT @32K & 6.90 s & \textbf{4.03 s} & 4.49 s & 4.39 s \\
\bottomrule
\end{tabular}
\end{table}

Table~\ref{tab:main} is the headline. Three observations.

\textbf{All quantized recipes match BF16 task accuracy within seed
noise.} Every quantized model
sits within seed noise of BF16 on every task: the three quantized
recipes span 0.54 points on the 5-task average --- less than one AIME
problem (3.3 points) --- and the largest single-task gap (RadixArk's
AIME $-2.5$) is inside the BF16 model's own seed spread (83.3--93.3). \minima{} matches
BF16's AIME'25 score exactly (26/30 on all four seeds) with all of GDN
at 4~bits. Generation behavior is unchanged too: \minima{} does not
``think longer'' (mean AIME generation 14{,}531 vs.\ 14{,}532 tokens
for BF16) and hits the 32K cap slightly \emph{less} often.

\textbf{Quantizing the GDN block yields measurable efficiency gains.}
\minima{} is the only checkpoint whose GDN
block (5.5B parameters, ${\sim}23\%$ of decode weight bytes) is at
4~bits, and it shows: $2.9\times$ smaller than BF16 in VRAM and on
disk, 7--13\% smaller than either community NVFP4 build, the largest KV
budget (1.81M cacheable tokens on one card), and the fastest prefill
(TTFT $6.90\,\mathrm{s} \to 4.03\,\mathrm{s}$ at 32K; $+14$--$19\%$
prompt throughput at 8K over Unsloth/RadixArk, whose GDN/attention
GEMMs remain FP8). Decode is weight-bandwidth-bound and all three
quantized models land within 4\% (RadixArk leads by 2--4\%;
\S\ref{sec:limitations}).

\textbf{Perplexity is the honest residual.} PPL is the one metric that
orders the recipes: Unsloth $<$ RadixArk $<$ \minima{} at both context
lengths, as expected --- the community recipes simply quantize less of
the model and pay for the headroom in 1.3--2.7\,GiB of weights. \minima{}'s gap to BF16
is $+0.72$ at 4K but only $+0.49$ at 32K, it never reaches a task
score, and \S\ref{sec:mechanism} shows it is a short-context, per-token
effect that context \emph{washes out} --- the opposite of the
error-accumulation the community recipes guard against.

\section{Why GDN survives 4 bits}\label{sec:mechanism}

Table~\ref{tab:main} refutes the community's caution empirically; this
section explains it. We captured the real inputs of all 48 GDN layers
while the BF16 model read eight 32K-token documents, re-implemented one
GDN layer standalone (verified against the reference implementation to
$6\times10^{-3}$ median relative output difference, i.e.\ BF16
rounding), and used NVFP4 \emph{fake quantization} --- quantize,
dequantize, continue in high precision --- to inject exactly the 4-bit
rounding error and nothing else. Four experiments form a chain.

\subsection{The inputs are not the reason}\label{sec:exp1}

The simplest hypothesis is that GDN sees an easier input distribution
than attention. It does not: GDN's projections read the \emph{same}
residual stream as attention (Table~\ref{tab:stats}). GDN inputs carry
extreme outliers (median-layer $\max/\mathrm{RMS}$ 63.5, kurtosis
${\sim}1{,}560$, hot channels $100\times$ the median channel), and
10--32\% of 16-element blocks are dominated by a single value. Yet the
actual per-token A4 quantization error is uniform across every layer
role --- 7.5--9.2\% --- because block scaling confines each outlier to
its 15 neighbors. Weight error (10.5--11.9\%, near-Gaussian weights)
\emph{exceeds} activation error everywhere. Both are flat in position
over the 32K window. Robustness must therefore come from what the
layer \emph{does} with the error, not from clean data.

\begin{table}[t]
\centering
\caption{Activation and weight statistics at NVFP4 granularity
(4$\times$32K captured tokens; median over layers). ``1-hot blocks'' =
share of 16-element blocks with $\max/\mathrm{RMS} > 3$ (one value
holds $>56\%$ of block energy). A4/W4 = relative error of quantizing
the activation/weight tensor alone.}
\label{tab:stats}
\small
\begin{tabular}{lrrrrr}
\toprule
input of & $\max/\mathrm{RMS}$ & kurtosis & 1-hot blocks \% & A4 err \% & W4 err \% \\
\midrule
GDN qkv/z/a/b & 63.5 & 1{,}564 & 10.6 & 7.6 & 10.6--11.6 \\
GDN out\_proj & 298.1 & 609 & 32.1 & 9.0 & 10.8 \\
attention q/k/v & 71.5 & 1{,}809 & 13.9 & 7.5 & 10.5--11.0 \\
attention o\_proj & 81.2 & 39 & 23.6 & 9.2 & 10.7 \\
MLP gate/up & 47.8 & 100 & 5.3 & 9.2 & 11.8--11.9 \\
MLP down & 368.3 & 962 & 24.3 & 8.9 & 11.8 \\
\bottomrule
\end{tabular}
\end{table}

\subsection{The protected projections are the safest ones}\label{sec:exp2}

We replayed each captured layer with exactly one projection quantized
at a time (96 replays over layers and sequences, 8K tokens each) and
measured the effect on the layer output $y$
(Table~\ref{tab:replay}). The result inverts the community's
precision map: fully quantizing the gate projections $a$ and $b$ ---
the two tensors every public recipe keeps in BF16 --- moves $y$ by
only 2.1\% and 2.6\%, the two smallest effects, even though their own
GEMM errors are 11.0\% and 8.5\%. The squashing in
Eq.~\eqref{eq:gates} is the shield: a ${\sim}11\%$ pre-activation
error becomes a 7.5\% error on $1-\alpha$ and a 5.2\% error on
$\beta$, and the recurrence (\S\ref{sec:exp3}) tolerates both. The
error \minima{} actually carries comes from the three plain GEMMs ---
\texttt{out} (12.7\%), \texttt{qkv} (10.4\%), \texttt{z} (9.9\%).
Two further regularities: the five projections' errors are
statistically independent (single-projection $y$ errors combine in
quadrature to $19.4\%$ vs.\ the measured $19.2\%$ for all-at-once),
and weights carry more error than activations for every projection
(W4A16 $>$ W16A4), consistent with Table~\ref{tab:stats}. Nothing
grows along the sequence (first vs.\ last quarter: 19.5\% vs.\
19.7\%).

\begin{table}[t]
\centering
\caption{Per-projection NVFP4 sensitivity: one projection W4A4 at a
time. ``gemm'', gate, and $y$ columns: median relative errors in \%
over 96 (layer, sequence) replays of 8K tokens. State column: relS
plateau of the 32K lockstep runs of \S\ref{sec:exp3} (median over 5
layers); 0 = the projection does not touch the recurrence.
\texttt{ab} is the pair the community protects; \texttt{all} is the
full \minima{} GDN block.}
\label{tab:replay}
\small
\begin{tabular}{lrrrrr}
\toprule
variant & gemm & $1-\alpha$ & $\beta$ & state $S$ & output $y$ \\
\midrule
a:W4A4 & 11.0 & 7.5 & --- & 3.6 & \textbf{2.1} \\
b:W4A4 & 8.5 & --- & 5.2 & 3.2 & \textbf{2.6} \\
ab:W4A4 & --- & 7.5 & 5.2 & 5.2 & 3.6 \\
qkv:W4A4 & 10.6 & --- & --- & 12.1 & 10.4 \\
z:W4A4 & 8.3 & --- & --- & 0 & 9.9 \\
out:W4A4 & 12.7 & --- & --- & 0 & 12.7 \\
all:W4A4 & --- & 7.5 & 5.2 & 12.6 & 19.2 \\
\bottomrule
\end{tabular}
\end{table}

\subsection{The recurrence bounds and erases the noise}\label{sec:exp3}

\begin{figure}[t]
\centering
\resizebox{\linewidth}{!}{\input{figs/impulse_decay}}
\caption{State error of the FP32 lockstep recurrence over 32K tokens,
five layers spread over depth (model layer indices).
\textbf{(a)} With the full \minima{} quantization injected at every
step, $\relS(t)$ plateaus immediately and stays flat --- no
accumulation. \textbf{(b)} A single 1\% state perturbation at
$t_0{=}1{,}024$ (log scale, 512-token bin means): the recurrence
erases it within a few hundred to a few thousand steps, orders of
magnitude faster than the decay-implied horizons of up to 62K tokens,
because the delta rule overwrites the state along each new key.}
\label{fig:propagation}
\end{figure}
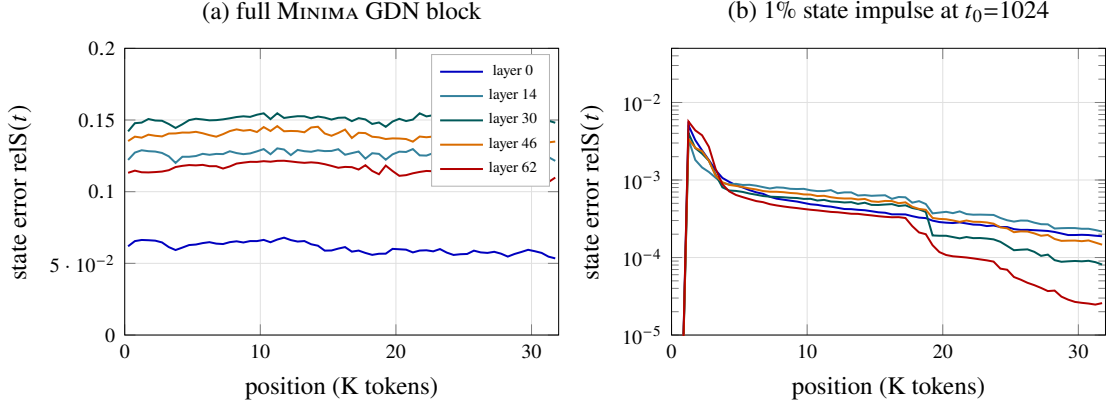

Does the 12.6\% state error of Table~\ref{tab:replay} grow over a long
context? We ran the recurrence in lockstep FP32 --- one clean
trajectory, eleven perturbed ones on identical inputs --- for 32K
tokens on five layers spread over depth. The full-\minima{} state
error is \emph{flat}: $\relS = 12.96\%$ at token 256 and $12.31\%$ at
token 32{,}768 (plateau 12.6\%, max 14.9\%;
Figure~\ref{fig:propagation}a). The recurrence reaches an equilibrium
where forgetting balances injection, and holds it for the entire
window.

The forgetting is faster than the decay gates alone explain. A one-off
1\% state impulse injected at $t{=}1{,}024$ falls to $1/e$ within
80--1{,}382 steps and to $1/10$ within ${\sim}2{,}200$--$2{,}900$ ---
while the decay-implied horizons $1/(1-\alpha)$ of the same layers
reach 44K--62K tokens. The extra erasure is the delta rule itself:
by Eq.~\eqref{eq:recurrence} every write \emph{overwrites} the state
along the current key direction, so old errors are deleted key by key
as new tokens arrive, not merely decayed.

A synthetic-noise arm locates the actual fragility, and thereby the
role of the parameterization. The state is very sensitive to
\emph{relative} noise applied directly to $\alpha$: $0.1\%$
multiplicative noise yields a 22\% state error, because with
$\alpha\approx1$ a tiny $\delta\alpha$ is an enormous relative change
in the horizon $1/(1-\alpha)$. Quantizing $a$ produces only 3.6\%
state error from an 11\% GEMM error because the noise lands on the
\emph{pre-activation} of Eq.~\eqref{eq:gates}, where softplus and the
exponential compress it before it touches the horizon. The log-space
gate parameterization --- chosen for training stability --- is
precisely what makes the gates quantization-proof at serving time.
Noise on $\beta$ is harmless outright (1\% noise $\to$ 0.4\% state
error): the delta rule's write is self-correcting, since a mis-scaled
correction is itself corrected by later writes.

\subsection{End to end, context washes the error out}\label{sec:exp4}

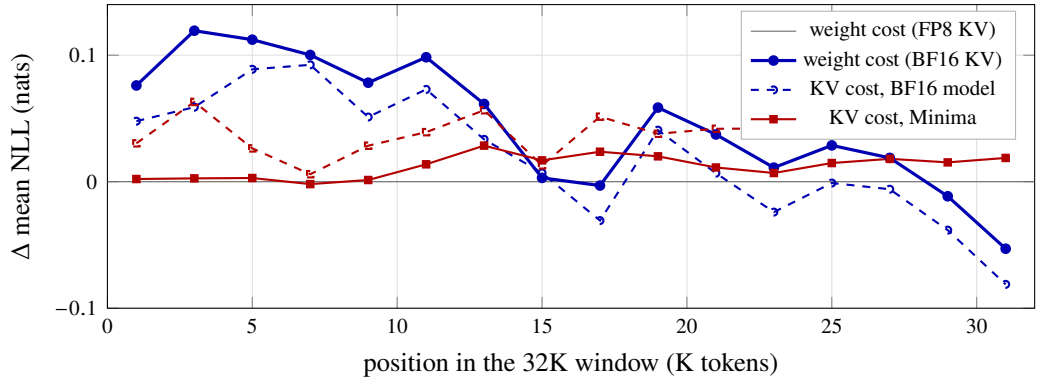
\begin{figure}[t]
\centering
\input{figs/position_deltas}
\caption{The two quantization costs, decomposed by position in the 32K
window (2K-token bins; full numbers in Table~\ref{tab:position}).
Blue: weight cost (\minima{}$-$BF16 at matched KV precision) ---
largest at the start, falling to zero and below by the end of the
window: the opposite of accumulation. Red: KV-cache cost (FP8$-$BF16
KV, same checkpoint) --- small, rising with position, larger for
\minima{}; this is the component the calibrated scales of
\S\ref{sec:kv} remove.}
\label{fig:position}
\end{figure}

If the mechanism above is right, the served model's quantization gap
should not grow with position --- and it should if the community's
accumulation picture were right. We split the per-token NLL of the 32K
perplexity runs by position (Figure~\ref{fig:position}). The
weight-quantization gap (\minima{}$-$BF16, same KV regime) is $+0.081$
nats in the first half of the window and $+0.011$ in the second; in
the final 2K tokens \minima{} scores \emph{better} than BF16
($-0.053$). The 4-bit cost is a short-context, per-token effect that a
filled state absorbs.

The FP8-KV cost behaves in exactly the opposite way: it is small,
rises with position, and is ${\sim}3\times$ larger for \minima{} ---
the signature of an attention-path effect rather than a weight effect.
\S\ref{sec:kv} eliminates it with calibrated scales.

\paragraph{Synthesis.}
Block scaling localizes the residual stream's outliers
(\S\ref{sec:exp1}); the gate nonlinearities compress what reaches the
control signals (\S\ref{sec:exp2}); the delta-rule recurrence bounds
the remaining noise at a plateau and actively erases it
(\S\ref{sec:exp3}); so the end-to-end cost shrinks with context
(\S\ref{sec:exp4}). None of this invokes luck or fine-tuning: it is
the architecture's own gating and correction structure. The
projections the community protects are exactly the ones the
architecture already protects.

\section{Measuring it correctly: serving-stack findings}\label{sec:systems}

Every conclusion above required first repairing the measurement
pipeline. We report four findings; each silently corrupted a result
before we caught it, and at least the first affects any hybrid-model
NVFP4 deployment today.

\paragraph{Per-module calibration vs.\ fused-GEMM scaling.}
llm-compressor calibrates one FP32 global scale per linear
\emph{module}; vLLM serves the GDN projections \emph{fused} ---
\texttt{in\_proj\_qkv}+\texttt{z} as one NVFP4 GEMM and
\texttt{in\_proj\_b}+\texttt{a} as another --- taking the maximum of
the constituent global scales without rescaling the local ones (the
ModelOpt path behaves the same). In our checkpoint the paired scales
differ by $1.82\times$ (qkv/z) and $2.75\times$ (b/a) in \emph{every}
one of the 48 layers, so the served model computed the decay and write
gates with mis-scaled weights. The corrupted model is deceptively
plausible: reasoning degrades moderately (AIME 80.8 vs.\ 86.7
repaired) while long-context perplexity gets \emph{better than BF16}
(a flat 6.86 at 32K vs.\ the true 10.84) --- a broken forget gate
makes the state hold everything, which happens to help next-token
prediction on WikiText. A checkpoint-side repair suffices: rewrite
each fused group to the shared global scale and fold the ratio into
the per-block E4M3 scales (94 scale sets across the model, worst ratio
$2.81\times$, re-rounding error $\le 6.2\%$ on affected local scales).
A GEMM-level probe confirms the fix (kernel-vs-reference error
$0.35/0.57 \to 0.002$), and every \minima{} number in this paper is
from the repaired checkpoint. The mismatch is invisible on checkpoints
that keep fused-adjacent modules at equal scales --- audits of the
Unsloth and RadixArk checkpoints found their fused groups uniform --- which is presumably why it has
gone unnoticed: it only bites recipes that quantize the GDN block,
which no public recipe had shipped at the time of our audit.

\paragraph{Composite vs.\ text-only serving path.}
The hub checkpoint is a multimodal composite; serving it makes vLLM
take a multimodal position-encoding path even for pure text, and that
path scores long context measurably differently (PPL@32K 10.04
composite vs.\ 10.22 text-only for the BF16 model) --- and both community
checkpoints are also composites. All models in this paper are served
from text-only extractions so that quantization is never confounded
with the serving path.

\paragraph{Raw-completion harnesses are invalid for thinking models.}
lm-eval's \texttt{local-completions} path sends few-shot prompts
without a chat template, so ``thinking disabled'' never reaches the
model: on MMLU-Pro it opened \texttt{<think>} on 25--48 of 50 sampled
questions and was truncated or self-interrupted, producing
per-subject swings of $\pm$40--60 points in \emph{both} directions
(invalid scores of 66.3/58.6 for BF16/\minima{} vs.\ the valid 80.4/79.7). All
short-tier numbers here use a chat-template harness with thinking
disabled and per-sample validity gates.

\paragraph{Context inversion in the base model.}
BF16 Qwen3.8-27B scores the \emph{same} tokens worse inside a 32K
request than in isolated 4K windows (PPL 6.95 $\to$ 10.35;
deterministic; reproduced identically in vLLM and in the reference
implementation to three decimals; retrieval at 64K remains 100\%).
This is a property of the model, not of quantization --- but it means
``PPL@32K'' comparisons are only meaningful within one serving path
and window protocol, which we hold fixed everywhere.

\section{KV-cache precision}\label{sec:kv}

The 48 GDN layers carry no KV cache; only the 16 attention layers do.
Storing their cache in FP8 (scale 1.0) is essentially free on tasks:
across BF16 and \minima{}, four seeds, six suites, no task score moves outside
seed spread, and capacity grows $1.8$--$1.9\times$. The one systematic
cost is perplexity at 32K: $+0.13$ for BF16 and $+0.41$ for \minima{}
--- $3\times$ larger, plausibly because \minima{}'s K/V projections
are already W4A4, leaving the values less headroom before the cache
rounds them again.

Calibrated scales close it. \minima{}+scales adds llm-compressor's
\texttt{kv\_cache\_scheme} (static per-tensor FP8 scales; 32 tensors on
16 attention layers --- exactly what Unsloth ships) to an otherwise
byte-identical \minima{} recipe. PPL@32K drops $10.84 \to 10.50$,
recovering 83\% of the penalty; the residual $+0.07$ is \emph{below}
BF16's own uncalibrated cost ($+0.13$). PPL@4K is unchanged, RULER
stays 100 at 32K/64K, and throughput matches \minima{} within 0.4\% on every
decode and prefill metric --- the scales are performance-free. The
practical recipe is therefore unconditional: quantize everything,
serve FP8 KV, ship calibrated scales.

\section{Related work}\label{sec:related}

\paragraph{Linear attention and hybrids.}
Gated DeltaNet \citep{yang2025gateddeltanet} combines the
parallelizable delta rule \citep{yang2024deltanet} with Mamba2-style
gating \citep{dao2024mamba2,gu2023mamba}; Qwen3.8-27B deploys it as
the dominant mixer in a hybrid stack. Our results speak to the
quantizability of this operator class, not to any one checkpoint's
training choices, since the mechanism (\S\ref{sec:mechanism}) rests on
the operator's own gating and correction structure.

\paragraph{Low-bit LLM quantization.}
Weight-only methods \citep{frantar2023gptq,lin2024awq} and
weight-activation methods \citep{xiao2023smoothquant,dettmers2022llmint8}
established that outlier handling is the central difficulty of W4/W8
inference; block-scaled microformats \citep{rouhani2023microscaling}
move the handling into the datatype, and NVFP4 \citep{nvidia2025nvfp4}
is the hardware-native instance we use. Prior work targets
transformer attention and MLPs; quantization of \emph{recurrent-state}
mixers in large hybrids has, to our knowledge, not been studied ---
the public recipes for this model simply exempt them
(\S\ref{sec:setup}). Concurrently with this work, QUASAR
\citep{counathe2026quasar} released a checkpoint of the same model that
quantizes all 496 projections, GDN included, via quantization-aware
training --- the 4-bit weights are learned by distillation from the
BF16
teacher.\footnote{\url{https://huggingface.co/QUASAR-QAT/Qwen3.8-27B-QUASAR-NVFP4}}
It appeared after our measurement campaign closed; its model card
reports a brief two-task spot check but no controlled study or
account of why the configuration survives; our results show the
training is not necessary ---
calibration-only PTQ reaches BF16-level task accuracy --- and
\S\ref{sec:mechanism} explains why. Concurrent engineering work in our group extends
the recipe studied here to the full model --- embeddings, the
language-model head, and the multi-token-prediction head also at NVFP4
W4A4 --- and further to 3-bit MLP codebooks with a distillation-healed
low-rank residual; notably, the sub-4-bit recipe keeps the GDN and
attention projections sealed at 4-bit weights, treating the result
established here as its foundation. That variant loads just
12.3\,GiB of weights into VRAM --- $4.1\times$ smaller than BF16's
50.1\,GiB, versus \minima{}'s 17.5\,GiB --- while
sustaining over 1{,}200 output tokens/s at 128-way concurrency on the
same single-GPU class; its accuracy evaluation uses different
protocols and is outside this paper's scope.

\paragraph{KV-cache compression.}
Post-training KV quantization \citep{hooper2024kvquant,liu2024kivi}
targets the dominant memory consumer of long-context attention
serving. Hybrids shrink that consumer architecturally (here only 16 of
64 layers cache KV); our contribution is the interaction term ---
FP8 KV costs a W4A4 model $3\times$ more perplexity than a BF16 model
unless calibrated scales are shipped, after which the cost is below
the BF16 model's own (\S\ref{sec:kv}).

\section{Limitations}\label{sec:limitations}

\textbf{Scope of evidence.} One model family and size (Qwen3.8-27B),
one quantization format (NVFP4), evaluated to 32K-token perplexity and
64K retrieval. We did not run longer-context stress tests: the
mechanism study answers the accumulation question directly (state
error flat over 32K, gap shrinking with position, retrieval perfect at
64K, ${\sim}14.5$K-token generations matching BF16 token-for-token),
and the bounded-error mechanism predicts longer contexts, but 128K+
behavior is extrapolation. The concurrent QAT checkpoint of
\citet{counathe2026quasar} is acknowledged but not benchmarked: it was
released after our measurement campaign closed, and it belongs to a
different recipe class (weights learned under quantization rather than
rounded post hoc), so its card's scores are not comparable to numbers
from our harness. \textbf{\minima{}+scales task scores} are
inherited from \minima{} rather than re-measured, justified by the KV
ablation showing no task movement for either model; PPL, RULER, and
throughput were re-measured. \textbf{Decode overhead.} \minima{}
trails RadixArk by 2--4\%
on decode despite fewer weight bytes; profiling attributes this to
small-batch NVFP4 activation-quantization overhead, a kernel-level
artifact rather than a property of the recipe. \textbf{Gates on other
architectures.} The gate-shielding argument (\S\ref{sec:exp3}) depends
on the log-space softplus/exponential parameterization; recurrent
mixers with linearly-parameterized decay may not enjoy it, and
$0.1\%$ direct noise on $\alpha$ demonstrably harms the state.

\section{Conclusion}\label{sec:conclusion}

We built a fully W4A4 NVFP4 version of a large hybrid LLM by
post-training quantization alone ---
all 496 backbone linear layers, Gated DeltaNet included --- and found it matches
BF16 across reasoning, knowledge, code, retrieval, and long-context
evaluation, while being the smallest and fastest-prefill recipe in its
cohort. The mechanism study explains the result architecturally: NVFP4
block scaling neutralizes the residual stream's outliers, the gate
parameterization compresses quantization noise before it reaches the
recurrence's control signals, and the delta rule's overwrite actively
erases state error faster than the decay gates alone would. The
community's protective precision maps guard exactly the projections
the architecture already guards. With the serving pipeline measured
correctly --- harmonized fused scales, text-only path, valid harness,
calibrated KV scales --- the practical guidance for hybrid models is
short: the recurrent half is the easy half; quantize it.

\bibliographystyle{plainnat}
\bibliography{refs}

\appendix

\section{Kernel-vs-reference numerics}\label{app:numerics}

The mechanism study's instrumented recurrence implements
Eq.~\eqref{eq:recurrence} in pure PyTorch at FP32; the served model
uses the fused chunkwise kernel. On identical 32K-token inputs, an FP32 kernel
run matches the reference to $7.7\times10^{-4}$ median relative output
difference --- the kernel's TF32 compute floor (its matmuls fix
\texttt{input\_precision='tf32'}) --- flat in position, with final
states agreeing to $4.4\times10^{-4}$. The BF16 kernel, as served,
sits at its rounding floor: $3.8\times10^{-3} \approx 2^{-8}$ median.
Because that floor exceeds the smallest perturbation the study
measures ($10^{-3}$), all error-propagation comparisons in
\S\ref{sec:exp3} are within-path: clean and perturbed trajectories
both run in the FP32 reference. One reproducibility note: the
reference must apply the kernel's default $1/\sqrt{K}$ query scale ---
omitting it produces a spurious $91.2\%$
($=1-1/\sqrt{128}$) output difference while final states still agree.

\section{Additional tables}\label{app:tables}

\subsection{Perplexity gap by position}\label{app:position}

Table~\ref{tab:position} decomposes the 32K WikiText-2 NLL by 2K-token
position bins, separating the weight-quantization gap
(\minima{}$-$BF16, same KV dtype) from the KV-compression gap
(FP8$-$BF16 KV, same checkpoint). Absolute NLL rises with position for
every model, BF16 included (context inversion, \S\ref{sec:systems});
the $\Delta$ columns are the signal. The weight gap falls from
$+0.081$ (first half, FP8 KV) to $+0.011$ (second half) and turns
negative in the last bins; the KV gap is small and \emph{rises} with
position, and is larger for \minima{} at every bin --- the pattern
that motivated the calibrated scales of \S\ref{sec:kv}.

\begin{table}[h]
\centering
\caption{Mean NLL (nats) by position in the 32K window. Anchor: the
BF16 model under FP8 KV. $\Delta$W = \minima{}$-$BF16 (weight cost, same KV);
$\Delta$KV = FP8$-$BF16 KV (same checkpoint). $0.01$ nats
$\approx 1\%$ perplexity.}
\label{tab:position}
\small
\begin{tabular}{lrrrrr}
\toprule
position & BF16 NLL & $\Delta$W (fp8 KV) & $\Delta$W (bf16 KV) & $\Delta$KV BF16 & $\Delta$KV \minima{} \\
\midrule
0--2K & 1.925 & $+0.076$ & $+0.048$ & $+0.002$ & $+0.030$ \\
2--4K & 1.993 & $+0.119$ & $+0.059$ & $+0.003$ & $+0.063$ \\
4--6K & 2.150 & $+0.112$ & $+0.089$ & $+0.003$ & $+0.026$ \\
6--8K & 2.074 & $+0.100$ & $+0.092$ & $-0.002$ & $+0.006$ \\
8--10K & 2.023 & $+0.078$ & $+0.051$ & $+0.001$ & $+0.029$ \\
10--12K & 2.117 & $+0.098$ & $+0.073$ & $+0.014$ & $+0.039$ \\
12--14K & 2.110 & $+0.062$ & $+0.034$ & $+0.029$ & $+0.057$ \\
14--16K & 2.146 & $+0.003$ & $+0.007$ & $+0.017$ & $+0.013$ \\
16--18K & 2.409 & $-0.003$ & $-0.031$ & $+0.024$ & $+0.051$ \\
18--20K & 2.456 & $+0.059$ & $+0.041$ & $+0.020$ & $+0.038$ \\
20--22K & 2.666 & $+0.037$ & $+0.007$ & $+0.011$ & $+0.042$ \\
22--24K & 2.739 & $+0.011$ & $-0.024$ & $+0.007$ & $+0.042$ \\
24--26K & 2.697 & $+0.029$ & $-0.001$ & $+0.015$ & $+0.045$ \\
26--28K & 2.575 & $+0.019$ & $-0.006$ & $+0.018$ & $+0.043$ \\
28--30K & 2.612 & $-0.012$ & $-0.038$ & $+0.015$ & $+0.042$ \\
30--32K & 2.700 & $-0.053$ & $-0.081$ & $+0.019$ & $+0.047$ \\
\midrule
first-half mean & & $+0.081$ & $+0.057$ & $+0.008$ & $+0.033$ \\
second-half mean & & $+0.011$ & $-0.017$ & $+0.016$ & $+0.044$ \\
\bottomrule
\end{tabular}
\end{table}

\subsection{Throughput}\label{app:throughput}

\begin{table}[h]
\centering
\caption{Full throughput sweep (one RTX~PRO~6000, TP=1, FP8 KV, same
serving config as every accuracy number). Decode: 1024-token prompts,
1024 generated tokens, concurrency 1/8/32. Prefill: single request,
one output token. \minima{}+scales differs from \minima{} only by the
32 KV scale tensors.}
\label{tab:throughput}
\small
\setlength{\tabcolsep}{4pt}
\begin{tabular}{lrrrrrrrr}
\toprule
 & \multicolumn{3}{c}{decode tok/s} & \multicolumn{3}{c}{TPOT ms} & \multicolumn{2}{c}{TTFT s} \\
model & @1 & @8 & @32 & @1 & @8 & @32 & @8K & @32K \\
\midrule
BF16 & 26 & 196 & 621 & 37.7 & 39.8 & 49.5 & 1.34 & 6.90 \\
\minima{} & 47 & 354 & 1{,}154 & 21.2 & 22.2 & 26.9 & 0.61 & 4.03 \\
\minima{}+scales & 47 & 355 & 1{,}152 & 21.2 & 22.2 & 27.0 & 0.61 & 4.04 \\
Unsloth & 49 & 364 & 1{,}132 & 20.2 & 21.6 & 27.3 & 0.72 & 4.49 \\
RadixArk & 51 & 376 & 1{,}174 & 19.6 & 20.9 & 26.3 & 0.69 & 4.39 \\
\bottomrule
\end{tabular}
\end{table}

\subsection{Error propagation, full variant table}\label{app:propagation}

Table~\ref{tab:propagation} gives the complete lockstep results behind
\S\ref{sec:exp3} (median over layers 0/14/30/46/62, 32K tokens).
Impulse decay per layer: $1/e$ after 550 / 164 / 281 / 80 / 1{,}382
steps; $1/10$ after 2{,}937 / 2{,}233 / 2{,}270 / 2{,}239 / 2{,}693.
Gate context:
mean $\alpha = 0.862$; per-layer mean horizons $1/(1-\alpha)$ of
43{,}970 / 1{,}895 / 4{,}156 / 8{,}342 / 61{,}659 tokens; mean
$\beta = 0.447$.

\begin{table}[h]
\centering
\caption{State error $\relS(t)$ vs.\ the clean FP32 recurrence
(\%, median over 5 layers) and resulting output errors.
alpha/beta-noise rows apply synthetic multiplicative noise directly to
the gates; impulse perturbs the state once at $t_0{=}1{,}024$ by 1\%.}
\label{tab:propagation}
\small
\setlength{\tabcolsep}{4pt}
\begin{tabular}{lrrrrrr}
\toprule
variant & $\relS$@256 & $\relS$@4096 & $\relS$@32768 & plateau & max & $y$ median \\
\midrule
minima (all W4A4) & 12.96 & 12.09 & 12.31 & 12.62 & 14.94 & 17.32 \\
qkv & 12.51 & 12.03 & 12.26 & 12.14 & 14.84 & 9.73 \\
z & 0 & 0 & 0 & 0 & 0 & 8.89 \\
a & 3.11 & 3.57 & 4.20 & 3.55 & 6.32 & 2.08 \\
b & 3.01 & 3.15 & 3.10 & 3.18 & 5.48 & 2.55 \\
out & 0 & 0 & 0 & 0 & 0 & 12.70 \\
ab & 4.55 & 5.21 & 5.26 & 5.15 & 8.19 & 3.56 \\
$\alpha$-noise 0.1\% & 2.33 & 12.73 & 19.65 & 22.17 & 40.93 & 9.50 \\
$\alpha$-noise 1\% & 21.97 & 40.94 & 42.83 & 46.06 & 67.29 & 25.63 \\
$\beta$-noise 1\% & 0.42 & 0.39 & 0.40 & 0.39 & 1.34 & 0.46 \\
impulse 1\% & 0 & 0.10 & 0.01 & 0.02 & 1.00 & 0.10 \\
\bottomrule
\end{tabular}
\end{table}

\subsection{Precision maps and statistics notes}\label{app:maps}

\minima{} quantizes 496 linear tensors to NVFP4 W4A4: per GDN layer
\texttt{in\_proj\_qkv}, \texttt{z}, \texttt{a}, \texttt{b},
\texttt{out\_proj} ($5\times48$); per attention layer
\texttt{q,k,v,o\_proj} ($4\times16$); per layer
\texttt{gate,up,down\_proj} ($3\times64$). Kept in BF16: embeddings,
\texttt{lm\_head}, the GDN \texttt{conv1d}, all norms,
$A_{\log}$/\texttt{dt\_bias}. Unsloth keeps GDN and attention at FP8
W8A8 with $a$/$b$ in BF16, NVFP4 in MLPs except layers 56--63 (FP8),
\texttt{lm\_head} FP8; RadixArk likewise protects GDN/attention, with
NVFP4 MLPs and \texttt{lm\_head}. Seed statistics for
Table~\ref{tab:main}: AIME'25 95\% CIs --- BF16 $[79.2, 94.2]$,
Unsloth $[82.4, 92.6]$, RadixArk $[81.5, 86.8]$; \minima{}'s four
seeds all scored 26/30, so its interval is degenerate and the per-seed
scores are the honest statement. GPQA-D: BF16 $[83.1, 89.9]$,
\minima{} $[82.8, 87.4]$, Unsloth $[82.4, 87.5]$, RadixArk
$[84.2, 86.5]$. Truncation at the 32K cap (AIME / GPQA): BF16
16.7 / 14.6\%, \minima{} 15.8 / 14.6\%, Unsloth 16.7 / 15.0\%,
RadixArk 19.2 / 14.4\% --- RadixArk's higher cap-hit rate is the likeliest cause
of its AIME dip. Zero request errors in every run.

\end{document}

%% file: figs/impulse_decay.tex
\begin{tikzpicture}
\begin{axis}[
  width=0.52\linewidth, height=5.6cm,
  title={(a) full \textsc{Minima} GDN block}, title style={font=\small},
  xlabel={position (K tokens)},
  ylabel={state error $\mathrm{relS}(t)$},
  xmin=0, xmax=32, ymin=0, ymax=0.2,
  ymode=normal,
  grid=major, grid style={black!12},
  tick label style={font=\scriptsize}, label style={font=\small},
  legend style={font=\tiny, draw=black!30},
]
\addplot[layer0, thick, no marks] coordinates { (0.25,0.0619106) (0.75,0.0655179) (1.25,0.0663164) (1.75,0.0661319) (2.25,0.0658393) (2.75,0.0646049) (3.25,0.0613516) (3.75,0.0592906) (4.25,0.0609949) (4.75,0.0627699) (5.25,0.0631326) (5.75,0.0641383) (6.25,0.0647797) (6.75,0.0637635) (7.25,0.0634704) (7.75,0.0647959) (8.25,0.0653741) (8.75,0.064556) (9.25,0.0655458) (9.75,0.0655235) (10.25,0.0663716) (10.75,0.0647222) (11.25,0.0666067) (11.75,0.0679277) (12.25,0.0662348) (12.75,0.0652167) (13.25,0.0655612) (13.75,0.0633192) (14.25,0.0623086) (14.75,0.0588781) (15.25,0.0589831) (15.75,0.0603017) (16.25,0.0613478) (16.75,0.0581711) (17.25,0.0589113) (17.75,0.0572607) (18.25,0.0559238) (18.75,0.0566959) (19.25,0.0567552) (19.75,0.0600973) (20.25,0.0598277) (20.75,0.0595543) (21.25,0.0576311) (21.75,0.0589279) (22.25,0.0591168) (22.75,0.0583735) (23.25,0.0599643) (23.75,0.0575036) (24.25,0.0558665) (24.75,0.0563646) (25.25,0.0565425) (25.75,0.0588674) (26.25,0.0576467) (26.75,0.0572904) (27.25,0.0580581) (27.75,0.0567627) (28.25,0.054726) (28.75,0.0565465) (29.25,0.057934) (29.75,0.0594548) (30.25,0.0584538) (30.75,0.0572685) (31.25,0.0546826) (31.75,0.0534751) };
\addlegendentry{layer 0}
\addplot[layer1, thick, no marks] coordinates { (0.25,0.122161) (0.75,0.127276) (1.25,0.128889) (1.75,0.128241) (2.25,0.127991) (2.75,0.127012) (3.25,0.124654) (3.75,0.120066) (4.25,0.124354) (4.75,0.124359) (5.25,0.124863) (5.75,0.126401) (6.25,0.125726) (6.75,0.126342) (7.25,0.126218) (7.75,0.128196) (8.25,0.126927) (8.75,0.126176) (9.25,0.126674) (9.75,0.128121) (10.25,0.129726) (10.75,0.126132) (11.25,0.130207) (11.75,0.129415) (12.25,0.12551) (12.75,0.129121) (13.25,0.12824) (13.75,0.127841) (14.25,0.128864) (14.75,0.12831) (15.25,0.128415) (15.75,0.129638) (16.25,0.125754) (16.75,0.126891) (17.25,0.128005) (17.75,0.123015) (18.25,0.126833) (18.75,0.126362) (19.25,0.122299) (19.75,0.12605) (20.25,0.127917) (20.75,0.127696) (21.25,0.129724) (21.75,0.125336) (22.25,0.125172) (22.75,0.126884) (23.25,0.127096) (23.75,0.128268) (24.25,0.126338) (24.75,0.123738) (25.25,0.121855) (25.75,0.124767) (26.25,0.125717) (26.75,0.123957) (27.25,0.125472) (27.75,0.12863) (28.25,0.121599) (28.75,0.125329) (29.25,0.12877) (29.75,0.129729) (30.25,0.129813) (30.75,0.128664) (31.25,0.124111) (31.75,0.121486) };
\addlegendentry{layer 14}
\addplot[layer2, thick, no marks] coordinates { (0.25,0.142074) (0.75,0.147776) (1.25,0.148191) (1.75,0.150696) (2.25,0.149641) (2.75,0.149469) (3.25,0.147202) (3.75,0.144433) (4.25,0.146986) (4.75,0.149902) (5.25,0.150876) (5.75,0.150384) (6.25,0.149762) (6.75,0.149921) (7.25,0.150443) (7.75,0.151201) (8.25,0.152079) (8.75,0.152575) (9.25,0.152177) (9.75,0.153675) (10.25,0.154667) (10.75,0.150612) (11.25,0.15469) (11.75,0.151415) (12.25,0.152107) (12.75,0.152641) (13.25,0.152582) (13.75,0.151204) (14.25,0.152405) (14.75,0.151654) (15.25,0.150315) (15.75,0.151543) (16.25,0.151707) (16.75,0.149191) (17.25,0.151298) (17.75,0.150395) (18.25,0.147238) (18.75,0.148265) (19.25,0.145236) (19.75,0.14891) (20.25,0.150842) (20.75,0.149267) (21.25,0.150296) (21.75,0.154654) (22.25,0.152516) (22.75,0.153348) (23.25,0.150568) (23.75,0.15396) (24.25,0.149593) (24.75,0.144967) (25.25,0.145738) (25.75,0.147448) (26.25,0.147605) (26.75,0.146722) (27.25,0.147153) (27.75,0.148597) (28.25,0.147208) (28.75,0.148904) (29.25,0.148189) (29.75,0.149394) (30.25,0.150942) (30.75,0.151944) (31.25,0.149518) (31.75,0.148076) };
\addlegendentry{layer 30}
\addplot[layer3, thick, no marks] coordinates { (0.25,0.135326) (0.75,0.138388) (1.25,0.137672) (1.75,0.139767) (2.25,0.138998) (2.75,0.138549) (3.25,0.140537) (3.75,0.140638) (4.25,0.141246) (4.75,0.141192) (5.25,0.141757) (5.75,0.140755) (6.25,0.139423) (6.75,0.138361) (7.25,0.140128) (7.75,0.142264) (8.25,0.143455) (8.75,0.142557) (9.25,0.143062) (9.75,0.141609) (10.25,0.14507) (10.75,0.142317) (11.25,0.145739) (11.75,0.142136) (12.25,0.142359) (12.75,0.142221) (13.25,0.141433) (13.75,0.144789) (14.25,0.145298) (14.75,0.140586) (15.25,0.139001) (15.75,0.140965) (16.25,0.143325) (16.75,0.137431) (17.25,0.141081) (17.75,0.140861) (18.25,0.138329) (18.75,0.137891) (19.25,0.136734) (19.75,0.137176) (20.25,0.137072) (20.75,0.136721) (21.25,0.134887) (21.75,0.138465) (22.25,0.137966) (22.75,0.138465) (23.25,0.137804) (23.75,0.136983) (24.25,0.134877) (24.75,0.135471) (25.25,0.13135) (25.75,0.132825) (26.25,0.134063) (26.75,0.13583) (27.25,0.136109) (27.75,0.136882) (28.25,0.135615) (28.75,0.135611) (29.25,0.136561) (29.75,0.13644) (30.25,0.134732) (30.75,0.134856) (31.25,0.134405) (31.75,0.134963) };
\addlegendentry{layer 46}
\addplot[layer4, thick, no marks] coordinates { (0.25,0.113168) (0.75,0.114671) (1.25,0.113542) (1.75,0.113427) (2.25,0.113676) (2.75,0.114042) (3.25,0.114979) (3.75,0.11719) (4.25,0.118017) (4.75,0.118565) (5.25,0.118346) (5.75,0.118694) (6.25,0.117773) (6.75,0.117787) (7.25,0.116143) (7.75,0.117609) (8.25,0.119598) (8.75,0.120261) (9.25,0.120829) (9.75,0.120772) (10.25,0.121279) (10.75,0.120908) (11.25,0.121503) (11.75,0.121578) (12.25,0.121077) (12.75,0.120519) (13.25,0.120116) (13.75,0.120463) (14.25,0.120082) (14.75,0.119182) (15.25,0.117986) (15.75,0.11668) (16.25,0.117284) (16.75,0.117104) (17.25,0.115441) (17.75,0.118448) (18.25,0.11448) (18.75,0.112166) (19.25,0.118197) (19.75,0.114419) (20.25,0.111062) (20.75,0.111535) (21.25,0.112829) (21.75,0.113769) (22.25,0.114457) (22.75,0.114254) (23.25,0.113078) (23.75,0.113861) (24.25,0.117027) (24.75,0.118702) (25.25,0.114921) (25.75,0.110189) (26.25,0.107833) (26.75,0.109304) (27.25,0.114075) (27.75,0.109622) (28.25,0.120682) (28.75,0.116994) (29.25,0.113011) (29.75,0.110663) (30.25,0.109888) (30.75,0.108856) (31.25,0.106688) (31.75,0.109835) };
\addlegendentry{layer 62}
\end{axis}
\begin{scope}[xshift=0.52\linewidth]
\begin{axis}[
  width=0.52\linewidth, height=5.6cm,
  title={(b) 1\% state impulse at $t_0{=}1024$}, title style={font=\small},
  xlabel={position (K tokens)},
  ylabel={state error $\mathrm{relS}(t)$},
  xmin=0, xmax=32, ymin=1e-5, ymax=0.05, legend pos=north east,
  ymode=log,
  grid=major, grid style={black!12},
  tick label style={font=\scriptsize}, label style={font=\small},
  legend style={font=\tiny, draw=black!30},
]
\addplot[layer0, thick, no marks] coordinates { (0.25,1e-06) (0.75,1e-06) (1.25,0.00507136) (1.75,0.00326791) (2.25,0.00241361) (2.75,0.00181035) (3.25,0.00134222) (3.75,0.00106382) (4.25,0.000946997) (4.75,0.000857444) (5.25,0.000800908) (5.75,0.000740308) (6.25,0.000693418) (6.75,0.000663541) (7.25,0.000608399) (7.75,0.000574053) (8.25,0.000557745) (8.75,0.000541511) (9.25,0.000527258) (9.75,0.000506062) (10.25,0.000486333) (10.75,0.000477889) (11.25,0.000461619) (11.75,0.000453271) (12.25,0.000437492) (12.75,0.000422332) (13.25,0.000420783) (13.75,0.000412505) (14.25,0.00040736) (14.75,0.000395598) (15.25,0.000383225) (15.75,0.000378105) (16.25,0.000360694) (16.75,0.00036088) (17.25,0.000359673) (17.75,0.000341677) (18.25,0.00032747) (18.75,0.000322338) (19.25,0.000299039) (19.75,0.000286802) (20.25,0.000281589) (20.75,0.000278451) (21.25,0.000283975) (21.75,0.00027507) (22.25,0.000266068) (22.75,0.000264377) (23.25,0.000254585) (23.75,0.000258966) (24.25,0.000247342) (24.75,0.000247151) (25.25,0.000232508) (25.75,0.000227245) (26.25,0.000227534) (26.75,0.000225166) (27.25,0.000222749) (27.75,0.000220088) (28.25,0.000213497) (28.75,0.00020416) (29.25,0.000195006) (29.75,0.000195101) (30.25,0.000196426) (30.75,0.000195101) (31.25,0.000191166) (31.75,0.000188171) };
\addplot[layer1, thick, no marks] coordinates { (0.25,1e-06) (0.75,1e-06) (1.25,0.00332868) (1.75,0.00181417) (2.25,0.0014514) (2.75,0.00127007) (3.25,0.00106969) (3.75,0.000912478) (4.25,0.000879185) (4.75,0.000892068) (5.25,0.000862877) (5.75,0.000867505) (6.25,0.00084434) (6.75,0.000809434) (7.25,0.000777634) (7.75,0.000803538) (8.25,0.000795382) (8.75,0.000770116) (9.25,0.000764928) (9.75,0.00076904) (10.25,0.000736867) (10.75,0.00071974) (11.25,0.000719985) (11.75,0.000740737) (12.25,0.000674939) (12.75,0.000692671) (13.25,0.000692522) (13.75,0.000631224) (14.25,0.000638559) (14.75,0.000625248) (15.25,0.000631236) (15.75,0.000635281) (16.25,0.000571181) (16.75,0.000603742) (17.25,0.00059803) (17.75,0.00050972) (18.25,0.000533809) (18.75,0.000489888) (19.25,0.000371031) (19.75,0.000379912) (20.25,0.00038862) (20.75,0.000373697) (21.25,0.000393545) (21.75,0.000363756) (22.25,0.000356739) (22.75,0.00035781) (23.25,0.000355964) (23.75,0.00035412) (24.25,0.000323683) (24.75,0.000302536) (25.25,0.00029079) (25.75,0.000300025) (26.25,0.000299048) (26.75,0.000281766) (27.25,0.00026926) (27.75,0.000267967) (28.25,0.000236691) (28.75,0.000239042) (29.25,0.000240199) (29.75,0.000239801) (30.25,0.000234851) (30.75,0.000235009) (31.25,0.000226407) (31.75,0.00021631) };
\addplot[layer2, thick, no marks] coordinates { (0.25,1e-06) (0.75,1e-06) (1.25,0.00401854) (1.75,0.00258095) (2.25,0.00218788) (2.75,0.00177444) (3.25,0.00117404) (3.75,0.000804639) (4.25,0.000732167) (4.75,0.000723058) (5.25,0.000701558) (5.75,0.000668732) (6.25,0.000645947) (6.75,0.000621233) (7.25,0.00060971) (7.75,0.000604662) (8.25,0.000598947) (8.75,0.000589985) (9.25,0.000585761) (9.75,0.000570764) (10.25,0.000572193) (10.75,0.000543682) (11.25,0.000554002) (11.75,0.000528216) (12.25,0.000523003) (12.75,0.000512994) (13.25,0.000517849) (13.75,0.000491678) (14.25,0.000509218) (14.75,0.000477506) (15.25,0.00047611) (15.75,0.000482852) (16.25,0.000488635) (16.75,0.000462282) (17.25,0.000471501) (17.75,0.00043543) (18.25,0.000417899) (18.75,0.000387204) (19.25,0.000192713) (19.75,0.000190942) (20.25,0.000190868) (20.75,0.00018365) (21.25,0.000176728) (21.75,0.0001839) (22.25,0.00017795) (22.75,0.000178515) (23.25,0.000175641) (23.75,0.000171453) (24.25,0.000159495) (24.75,0.000141058) (25.25,0.000124217) (25.75,0.000126117) (26.25,0.000127354) (26.75,0.000118425) (27.25,0.000105037) (27.75,0.000109048) (28.25,9.24665e-05) (28.75,8.81049e-05) (29.25,8.93258e-05) (29.75,9.02652e-05) (30.25,8.93144e-05) (30.75,9.07175e-05) (31.25,8.71918e-05) (31.75,8.12961e-05) };
\addplot[layer3, thick, no marks] coordinates { (0.25,1e-06) (0.75,1e-06) (1.25,0.00335697) (1.75,0.0026028) (2.25,0.00235243) (2.75,0.00177999) (3.25,0.00113285) (3.75,0.000937094) (4.25,0.000864008) (4.75,0.000839481) (5.25,0.000811209) (5.75,0.000778341) (6.25,0.000755263) (6.75,0.000720641) (7.25,0.000706003) (7.75,0.000705144) (8.25,0.000696455) (8.75,0.000680497) (9.25,0.000672237) (9.75,0.000651855) (10.25,0.000648712) (10.75,0.000620971) (11.25,0.000625558) (11.75,0.000600022) (12.25,0.000583988) (12.75,0.000577215) (13.25,0.000578154) (13.75,0.000560645) (14.25,0.000569795) (14.75,0.000530334) (15.25,0.000522277) (15.75,0.00053733) (16.25,0.000534519) (16.75,0.000501623) (17.25,0.000515147) (17.75,0.000451732) (18.25,0.000423621) (18.75,0.000412804) (19.25,0.00032392) (19.75,0.000314893) (20.25,0.000311581) (20.75,0.000300711) (21.25,0.000288235) (21.75,0.000300377) (22.25,0.000289173) (22.75,0.000288346) (23.25,0.000283289) (23.75,0.000273229) (24.25,0.000244734) (24.75,0.000237986) (25.25,0.000212173) (25.75,0.000217392) (26.25,0.000216322) (26.75,0.000204873) (27.25,0.000190919) (27.75,0.000195312) (28.25,0.0001735) (28.75,0.000164046) (29.25,0.000165059) (29.75,0.000165466) (30.25,0.000162478) (30.75,0.000166079) (31.25,0.00015603) (31.75,0.000146487) };
\addplot[layer4, thick, no marks] coordinates { (0.25,1e-06) (0.75,1e-06) (1.25,0.00568164) (1.75,0.00436977) (2.25,0.00377801) (2.75,0.00268752) (3.25,0.00139029) (3.75,0.000881527) (4.25,0.000714111) (4.75,0.00064126) (5.25,0.000601269) (5.75,0.000566238) (6.25,0.00053461) (6.75,0.000517757) (7.25,0.000487421) (7.75,0.000469859) (8.25,0.000455803) (8.75,0.00044372) (9.25,0.000434297) (9.75,0.00042325) (10.25,0.000413309) (10.75,0.000405839) (11.25,0.000396942) (11.75,0.000387784) (12.25,0.000383744) (12.75,0.000375591) (13.25,0.00036694) (13.75,0.000365766) (14.25,0.000356806) (14.75,0.000349566) (15.25,0.000341275) (15.75,0.000335226) (16.25,0.000330726) (16.75,0.000331877) (17.25,0.000323268) (17.75,0.000254252) (18.25,0.000210663) (18.75,0.000199906) (19.25,0.000144057) (19.75,0.000117586) (20.25,0.000107726) (20.75,0.00010369) (21.25,0.000101923) (21.75,0.000100086) (22.25,9.74749e-05) (22.75,9.4645e-05) (23.25,9.1917e-05) (23.75,8.85435e-05) (24.25,7.18959e-05) (24.75,6.95481e-05) (25.25,5.60549e-05) (25.75,5.18085e-05) (26.25,4.71895e-05) (26.75,4.40121e-05) (27.25,4.05353e-05) (27.75,3.69177e-05) (28.25,3.74507e-05) (28.75,3.12973e-05) (29.25,2.85992e-05) (29.75,2.67562e-05) (30.25,2.60324e-05) (30.75,2.54819e-05) (31.25,2.47078e-05) (31.75,2.5807e-05) };
\end{axis}
\end{scope}
\end{tikzpicture}

%% file: figs/position_deltas.tex
\begin{tikzpicture}
\begin{axis}[
  width=0.94\linewidth, height=5.6cm,
  xlabel={position in the 32K window (K tokens)},
  ylabel={$\Delta$ mean NLL (nats)},
  xmin=0, xmax=32, ymin=-0.1, ymax=0.14,
  grid=major, grid style={black!12},
  legend style={font=\scriptsize, at={(0.98,0.98)}, anchor=north east, draw=black!30},
  tick label style={font=\scriptsize}, label style={font=\small},
]
\addplot[black!60, no marks, domain=0:32, samples=2] {0};
\addplot[col1, very thick, mark=*, mark size=1.4pt] coordinates { (1,0.0760865) (3,0.119357) (5,0.112284) (7,0.100192) (9,0.0782675) (11,0.0983767) (13,0.0614665) (15,0.00301935) (17,-0.00301997) (19,0.0585845) (21,0.0373417) (23,0.0111351) (25,0.0287094) (27,0.0188269) (29,-0.0114984) (31,-0.0529819) };
\addlegendentry{weight cost (FP8 KV)}
\addplot[col1, thick, dashed, mark=o, mark size=1.3pt] coordinates { (1,0.0478623) (3,0.0589305) (5,0.0888542) (7,0.0923816) (9,0.0511497) (11,0.0729616) (13,0.0335002) (15,0.0069679) (17,-0.0307204) (19,0.040756) (21,0.00675763) (23,-0.0240113) (25,-0.00103747) (27,-0.00594362) (29,-0.0382742) (31,-0.0811392) };
\addlegendentry{weight cost (BF16 KV)}
\addplot[col2, thick, mark=square*, mark size=1.2pt] coordinates { (1,0.00212213) (3,0.00263958) (5,0.00292118) (7,-0.00189559) (9,0.00134223) (11,0.0137511) (13,0.0285224) (15,0.0168718) (17,0.0237042) (19,0.0200638) (21,0.0112594) (23,0.0068856) (25,0.0147566) (27,0.0180744) (29,0.0152755) (31,0.01879) };
\addlegendentry{KV cost, BF16 model}
\addplot[col2, thick, dashed, mark=square, mark size=1.2pt] coordinates { (1,0.0303463) (3,0.0630662) (5,0.0263513) (7,0.00591511) (9,0.02846) (11,0.0391662) (13,0.0564888) (15,0.0129233) (17,0.0514046) (19,0.0378923) (21,0.0418435) (23,0.042032) (25,0.0445035) (27,0.042845) (29,0.0420514) (31,0.0469473) };
\addlegendentry{KV cost, Minima}
\end{axis}
\end{tikzpicture}